\documentclass{ifacconf}

\usepackage{graphicx}      
\usepackage{natbib}        

\usepackage{amsmath,amssymb,amsfonts}
\usepackage{array} 
\usepackage{microtype} 
\usepackage{balance} 
\usepackage{color} 
\usepackage{bm} 
\usepackage{tabularx} 
\usepackage{multirow} 
\usepackage{booktabs} 
\usepackage{threeparttable} 
\usepackage{comment} 
\usepackage{footnote}

\begin{document}
\begin{frontmatter}

\title{Bioinspired Composite Learning Control Under Discontinuous Friction for\\Industrial Robots\thanksref{footnoteinfo}}

\thanks[footnoteinfo]{This work was supported in part by the Guangdong Pearl River Talent Program of China under Grant No. 2019QN01X154 (\textit{Corresponding author: Yongping Pan}).}

\author[First]{Yongping Pan}
\author[Second]{Kai Guo}
\author[Third]{Tairen Sun}
\author[Fourth]{Mohamed Darouach}

\address[First]{School of Computer Science and Engineering, Sun Yat-sen University, Guangzhou 510006, China (e-mail: panyongp@mail.sysu.\-edu.cn).}
\address[Second]{School of Mechanical Engineering, Shandong Univ\-ersity, Jinan 250061, China (e-mail: kaiguo@sdu.edu.cn)}
\address[Third]{School of Medical Instrument and Food Engineering, University of Shanghai for Science and Technology, Shanghai 200093, China (e-mail: suntren@gmail.com)}
\address[Fourth]{CRAN UMR 7035 CNRS, University of Lorraine, Cosnes et Romain 54400, France (e-mail: mohamed.darouach@univ-lorraine.fr)}

\begin{abstract}                
Adaptive control can be applied to robotic systems with parameter uncertainties, but improving its performance is usually difficult, especially under discontinuous friction. Inspired by the human motor learning control mechanism, an adaptive learning control approach is proposed for a broad class of robotic systems with discontinuous friction, where a composite error learning technique that exploits data memory is employed to enhance parameter estimation. Compared with the classical feedback error learning control, the proposed approach can achieve superior transient and steady-state tracking without high-gain feedback and persistent excitation at the cost of extra computational burden and memory usage. The performance improvement of the proposed approach has been verified by experiments based on a DENSO industrial robot.
\end{abstract}


\end{frontmatter}

\section{Introduction}\label{Introduction}
%
%
%
%

The feedback error learning (FEL) framework is a computational model of human motor learning control in the cerebellum, where the limitations on feedback delays and small feedback gains in biological systems can be overcome by internal forward and inverse models, respectively \citep{Kawato1988}. The name ``FEL'' emphasizes the usage of a feedback control signal for the heterosynaptic learning of internal neural models. There are two key features for the FEL: Internal dynamics modeling and hybrid feedback-feedforward (HFF) control, which are well supported by much neuroscientific evidence \citep{Thoroughman2000, Wolpert1998, Morasso2005}.

A simplified FEL architecture without the internal forward model has been extensively studied for robot control \citep{Kawato1988, Tolu2012a, Gomi1993, Hamavand1995, Talebi1998, Topalov1998, Teshnehlab1996, Kalanovic2000, Kurosawa2005, Neto2010, Jo2011}. However, stability guarantee relies on a precondition that the controlled plant can be stabilized by linear feedback without feedforward control in \cite{Tolu2012a, Kawato1988, Teshnehlab1996, Kalanovic2000, Kurosawa2005, Neto2010, Jo2011}, which may not be satisfied for many control problems such as robot tracking control; in \cite{Gomi1993, Hamavand1995, Talebi1998, Topalov1998}, internal inverse models are implemented in the feedback loops, which violates the original motivation of proposing FEL. stability analysis of FEL control for a class of nonlinear systems was investigated in \cite{Nakanishi2004}, where the feedback gain is required to be sufficiently large to compensate for plant uncertainty so as to guarantee closed-loop stability. The approach of \cite{Nakanishi2004} was applied to the rehabilitation of Parkinson’s disease in \cite{Rouhollahi2017}. However, the accurate capture of the plant dynamics is not fully investigated and discontinuous friction is largely neglected in existing FEL robot control methods.

This paper proposes a bioinspired adaptive learning control approach for a broad class of robotic systems with discontinuous friction, where a composite error learning (CEL) technique exploiting data memory is applied to enhance parameter estimation. The word ``composite'' refers to the composite exploitation of instantaneous data and data memory, and the composite exploitation of the tracking error and a generalized predictive error for parameter learning.
Compared with the classical FEL control, the proposed approach can achieve superior transient and steady-state tracking without high-gain feedback and persistent excitation (PE). Exponential convergence of both the tracking error and the parameter estimation error is guaranteed under an interval-excitation (IE) condition that is much weaker than PE. 

In the HFF control, the use of the desired output as the regressor input leads to two attractive merits \citep{Sadegh1990}:
1) The regressor output can be calculated and stored offline to significantly reduce the amount of online calculations;
2) the noise correlation between the parameter estimation error in the parameter update law and the adaptation signal (i.e. an infinite gain phenomenon) can be removed to enhance estimation robustness.
Additional advantages of HFF control based on neural networks can be referred to \cite{Pan2016a, Pan2017a}. This study is based on our previous studies in composite learning control \citep{Pan2016b, Pan2016c, Pan2017b, Pan2018, Pan2019c, Guo2019, Guo2020, Guo2022a, Guo2022b}, in which the methods of \cite{Pan2016b, Pan2016c, Pan2018, Pan2019c} do not consider discontinuous friction, the methods of \cite{Pan2016b, Pan2016c, Pan2017b} consider only the case of $M(\bm q)$ in (\ref{eq01}) being a known constant, and all the above methods do not resort to the HFF scheme.

In the rest of this article, the control problem is formulated in Sec. \ref{Problem}; the CEL control is presented in Sec. \ref{Bioinspired}; experimental results are given in Sec. \ref{Example}; conclusions are drawn in Sec. \ref{Conclusions}. Throughout this paper, $\mathbb{R}$, $\mathbb{R}^+$, $\mathbb{R}^n$ and $\mathbb{R}^{n\times m}$ denote the spaces of real numbers, positive real numbers, real $n$-vectors and real $n\times m$-matrices, respectively, $\|\mathbf x\|$ denotes the Euclidian norm of $\mathbf x$, $L_\infty$ denotes the space of bounded signals, tr$(A)$ denotes the trace of $A$, $\lambda_{\min}(A)$ denotes the minimal eigenvalue of $A$, diag$(\cdot)$ denotes a diagonal matrix, $\min(\cdot)$, $\max(\cdot)$ and $\sup(\cdot)$ denote the operators of minimum, maximum and supremum, respectively, $\mathcal B_c$ $:=$ $\{\mathbf x| \|\mathbf x\| \leq c\}$ is the ball of radius $c$, and ${\mathcal{C}}^k$ represents the space of functions whose $k$-order derivatives all exist and are continuous, where $c \in \mathbb R^+$, $\mathbf x \in \mathbb R^n$, $A \in \mathbb R^{n\times n}$, and $n$, $m$ and $k$ are natural numbers. In the subsequent sections, for the sake of brevity, the argument(s) of a function may be omitted while the context is sufficiently explicit.

\section{Biological Background}\label{Background}

The FEL control process for human voluntary movements is described the following procedure \citep{Wolpert1998}:
\begin{enumerate}
\item The association cortex sends a desired output  $\bm q_\mathrm{d}$ in the body coordinates to the motor cortex;
\item A motor command $\bm u$ is computed in the motor cortex and is transmitted to muscles via spinal motoneurons to generate a control torque $\bm\tau$;
\item The musculoskeletal system realizes an actual movement $\bm q$ by interacting with its environment;
\item The $\bm q$ is measured by proprioceptors and is fed back to the motor cortex via the transcortical loop;
\item The spinocerebellum-magnocellular red nucleus system requires an internal neural model of the musculoskeletal system (i.e. internal forward model) while monitoring $\bm u$ and $\bm q$ to predict $\hat{\bm q}$, where the predictive error $\tilde{\bm q}$ is transmitted to the motor cortex via the ascending pathway and to muscles through the rubrospinal tract as the modification of $\bm u$;
\item The cerebrocerebellum-parvocellular red nucleus system requires an internal neural model for the inverse modeling of the musculoskeletal system (i.e. internal inverse model) while receiving the desired output $\bm q_\mathrm{d}$ and a feedback command $\bm u_\mathrm{FB}$;
\item As the motor learning proceeds, a feedforward command $\bm u_\mathrm{FF}$ generated by the internal inverse model gradually takes place of $\bm u_\mathrm{FB}$ as the main command;
\item Once the internal inverse model is learnt, it generates the motor command $\bm u$ directly using $\bm q_\mathrm{d}$ to perform various tasks precisely without external feedback.
\end{enumerate}

\begin{figure}[!t]
\centering
\includegraphics[width = 3.4in]{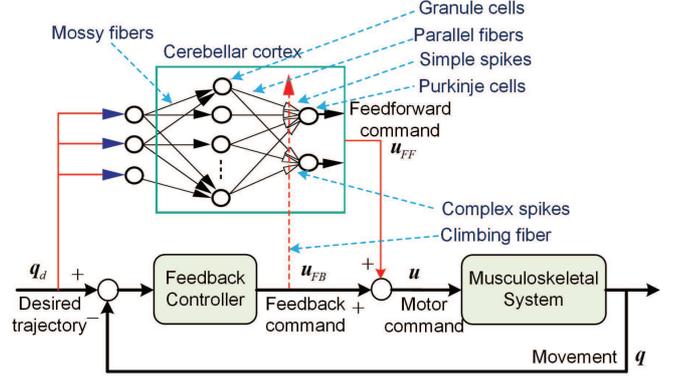}
\caption{A cerebellar neural circuit of simplified FEL control which is redrawn according to \cite{Wolpert1998}, where the internal forward model is omitted.}
\label{Fig02}
\end{figure}

The cerebellar neural circuit of a simplified FEL framework without the internal forward model is demonstrated in Fig. \ref{Fig02}, in which the simple spikes of Purkinje cells represent $\bm u_{\mathrm{FF}}$, the parallel fiber inputs receive  $\bm q_\mathrm{d}$, the climbing fiber inputs receive $\bm u_{\mathrm{FB}}$, and the complex spikes of Purkinje cells activated by the climbing fiber inputs represent sensory error signals in motor command coordinates.

\section{Problem Formulation}\label{Problem}

Consider a general class of robotic systems described by an Euler-Lagrange formulation \citep{Pan2016a}\footnotemark[1]:
\footnotetext[1]{It can be regarded as a simplified model of the musculoskeletal system where the muscle dynamics is ignored resulting in $\bm\tau = \bm u$.}
\begin{equation}\label{eq01}
M(\bm q)\ddot{\bm q} + C(\bm q,\dot{\bm q}) \dot{\bm q} + G(\bm q) + F(\dot{\bm q}) = \bm \tau
\end{equation}
where $\bm q(t)$ $= [q_1(t), q_2(t), \cdots, q_n(t)]^T \in \mathbb R^n$ is a joint angle vector, $M(\bm q) \in \mathbb R^{n\times n}$ is an inertia matrix, $C(\bm q$, $\dot{\bm q})$ $\in$ $\mathbb R^{n\times n}$ is a centripetal-Coriolis matrix, $G(\bm q) \in \mathbb R^n$, $F(\dot{\bm q}) \in \mathbb R^n$ and $\bm \tau(t)\in \mathbb R^n$ denote gravitational, friction, and control torque vectors, respectively, and $n$ is the number of links. It is assumed that $F(\dot{\bm q})$ can be expressed as follows:
\begin{equation}\label{eq02a}
F(\dot{\bm q}) = F_v\dot{\bm q} + F_c\mathrm{sgn}(\dot{\bm q})
\end{equation}
with $F_v :=$ diag$(k_{v1}, k_{v2}, \cdots, k_{vn})$, $F_c :=$ diag$(k_{c1}$, $k_{c2}$, $\cdots$, $k_{cn})$, and $\mathrm{sgn}(\dot{\bm q})$ $:=$ [$\mathrm{sgn}(\dot{q}_1)$, $\mathrm{sgn}(\dot{q}_2)$, $\cdots$, $\mathrm{sgn}(\dot{q}_n)]^T$, where $k_{vi} \in \mathbb R^+$ are coefficients of viscous friction, $k_{ci} \in \mathbb R^+$ are coefficients of Coulomb friction, and $i =$ 1 to $n$. For facilitating presentation, define a lumped uncertainty
\begin{equation}\label{eq02}
H(\bm q, \dot{\bm q}, \bm v, \dot{\bm v}) := M(\bm q)\dot{\bm v} + C(\bm q,\dot{\bm q})\bm v + G(\bm q)
\end{equation}
with ${\bm v} \in \mathbb R^n$ an auxiliary variable. The following properties from \cite{Pan2018} and definitions from \cite{Kingravi2012} are introduced to facilitate control design.

\textbf{Property 1}: $M(\bm q)$ is a symmetric and positive-definite matrix which satisfies $m_0 \|\bm\zeta\|^2 \leq \bm\zeta^T M(\bm q) \bm\zeta \leq \bar m\|\bm\zeta\|^2$, $\forall$ $\bm\zeta$ $\in \mathbb R^n$, in which $m_0$, $\bar m$ $\in \mathbb R^+$ are some constants.

\textbf{Property 2}: $\dot{M}(\bm q) - 2C(\bm q,\dot{\bm q})$ is skew-symmetric such that $\bm\zeta^T(\dot{M}(\bm q)$ $- 2C(\bm q,\dot{\bm q}))\bm\zeta = 0$, $\forall \bm\zeta \in \mathbb R^n$, implying the internal forces of the robot do no work.

\textbf{Property 3}: $F(\dot{\bm q})$ can be parameterized by
\begin{align}\label{eq05a}
F(\dot{\bm q}) = \Phi^T_f(\dot{\bm q})W_f & = \underbrace{\Phi^T_v(\dot{\bm q})W_v} + \underbrace{\Phi^T_c(\dot{\bm q})W_c}\\
& \;\;\;\;\;\;\;\;\; F_v\dot{\bm q}\;\;\;\;\;\;\;\;\; F_c\mathrm{sgn}(\dot{\bm q})\notag
\end{align}
where $\Phi_f := [\Phi_v^T, \Phi_c^T]^T \in \mathbb{R}^{2n\times n}$ is a known regressor, $W_f$ $:=$ $[W_v^T, W_c^T]^T \in \mathcal B_{c_f}$ $\subset$ $\mathbb{R}^{2n}$ is an unknown parameter vector, $c_f \in \mathbb R^+$ is a certain constant, and
\begin{gather*}
 W_v := [k_{v1}, k_{v2}, \cdots, k_{vn}]^T, \\
 W_c := [k_{c1}, k_{c2}, \cdots, k_{cn}]^T,\\
\Phi_v := \mathrm{diag}(\dot q_1, \dot q_2, \cdots, \dot q_n), \\
\Phi_c := \mathrm{diag}(\mathrm{sgn}(\dot q_1), \mathrm{sgn}(\dot q_2), \cdots, \mathrm{sgn}(\dot q_n)).
\end{gather*}

\textbf{Property 4}: $H(\bm q, \dot{\bm q}, \bm v, \dot{\bm v})$ can be parameterized by
\begin{equation}\label{eq03}
H(\bm q, \dot{\bm q}, \bm v, \dot{\bm v}) = \Phi^T_h(\bm q, \dot{\bm q}, \bm v, \dot{\bm v})W_h
\end{equation}
where $\Phi_h:\mathbb R^{4n}\mapsto\mathbb R^{N\times n}$ is a known $\mathcal C^1$ regressor, $W_h \in \mathcal B_{c_h}$ $\subset \mathbb R^{N}$ is an unknown parameter vector, $c_h \in \mathbb R^+$ is a certain constant, and $N$ is the dimension of $W_h$.

\textbf{Definition 1}: A bounded signal $\Phi(t) \in \mathbb R^{N\times n}$ is of IE if $\exists$ $T_e$, $\tau_d$, $\sigma_e \in \mathbb R^+$ such that $\int_{T_e-\tau_d}^{T_e}\Phi(\tau)\Phi^T(\tau)d\tau \geq \sigma_e I$.

\textbf{Definition 2}: A bounded signal $\Phi(t) \in \mathbb R^{N\times n}$ is of PE if $\exists$ $\sigma_e$, $\tau_d$ $\in \mathbb R^+$ such that $\int_{t-\tau_d}^{t}\Phi(\tau)\Phi^T(\tau)d\tau \geq \sigma_e I$, $\forall t \geq 0$.

Let $c_w := (c_h^2 + c_f^2)^{1/2}$. It follows from \eqref{eq02}-\eqref{eq03} that the left side \eqref{eq01} can be written as a parameterized form:
\begin{equation}\label{eq03b}
H(\bm q,\dot{\bm q},\dot{\bm q},\ddot{\bm q}) + F(\dot{\bm q}) = \Phi^T(\bm q,\dot{\bm q},\ddot{\bm q})W
\end{equation}
in which $\Phi(\bm q,\dot{\bm q},\ddot{\bm q})$ $:=$ $[\Phi_h^T(\bm q,\dot{\bm q},\dot{\bm q},\ddot{\bm q}), \Phi_v^T(\dot{\bm q}), \Phi_c^T(\dot{\bm q})]^T \in \mathbb{R}^{(N+2n)\times n}$ and $W := [W_h^T, W_v^T, W_c^T]^T \in \mathcal B_{c_w} \subset \mathbb{R}^{N+2n}$.
Let $\hat W_h(t)$ $\in \mathbb R^{N}$, $\hat W_f(t)$ $\in \mathbb R^{2n}$, $\hat W_v(t)$ $\in \mathbb R^{n}$ and $\hat W_c(t)$ $\in \mathbb R^{n}$ denote estimates of $W_h$, $W_f$, $W_v$ and $W_c$, respectively. Define a parameter estimation error $\tilde{W}$ $:= W - \hat W$ $=$ $[\tilde{W}_h^T$, $\tilde{W}_v^T$, $\tilde{W}_c^T]^T$, where $\hat W := [\hat W_h^T$, $\hat W_v^T$, $\hat W_c^T]^T \in$ $\mathbb{R}^{N+2n}$, $\tilde{W}_h$ $:= W_h - \hat W_h$, $\tilde{W}_v$ $:=$ $W_v$ $-$ $\hat W_v$, and $\tilde{W}_c$ $:=$ $W_c$ $-$ $\hat W_c$.

Let $\mathbf x_\mathrm{d}(t)$ $:=$ $[\bm q_\mathrm{d}^T(t)$, $\dot{\bm q}_\mathrm{d}^T(t)$, $\ddot{\bm q}_\mathrm{d}^T(t)]^T$ $\in$ $\mathbb R^{3n}$ be of $L_\infty$ with $\bm q_\mathrm{d} :=$ $[q_{\mathrm{d}1}$, $q_{\mathrm{d}2}$, $\cdots$, $q_{\mathrm{d}n}]^T$ $\in$ $\mathbb R^n$ a desired output. Define a position tracking error $\mathbf e_1 := \bm q_\mathrm{d} - \bm q$ and a ``reference velocity'' tracking error $\mathbf e_2 := \dot{\mathbf e}_1 + \Lambda_1\mathbf e_1$ $= \dot{\bm q}_\mathrm{r} - \dot{\bm q}$, where $\Lambda_1 \in \mathbb R^{n\times n}$ denotes a positive-definite diagonal matrix, and $\dot{\bm q}_\mathrm{r} := \dot{\bm q}_\mathrm{d} + \Lambda_1\mathbf e_1$ denotes a ``reference velocity''. Let $\mathbf e =$ $[\mathbf e_1^T, \mathbf e_2^T]^T$ $\in$ $\mathbb R^{2n}$. The objective of this study is to develop an adaptive control strategy for the system (\ref{eq01}), such that the closed-loop system is stable with guaranteed convergence of both $\mathbf e$ and $\tilde{W}$.

\section{Robot Control Design}\label{Bioinspired}

\subsection{Hybrid Feedback-Feedforward Structure}\label{Structure}

In this subsection, we have $\bm v$ $=$ $\dot{\bm q}_\mathrm{r}$ in \eqref{eq03}. Taking the time derivative of $\mathbf e_2$ and multiplying $M(\bm q)$, one obtains
\[M(\bm q) \dot{\mathbf e}_2 = M(\bm q) (\ddot{\bm q}_\mathrm{d} + \Lambda \dot{\mathbf e}_1) - M(\bm q) \ddot{\bm q}.\]
Substituting the expression of $M(\bm q) \ddot{\bm q}$ by (\ref{eq01}) into the foregoing equality, one gets the tracking error dynamics
\begin{align*}
M(\bm q) \dot{\mathbf e}_2 & = M(\bm q) \ddot{\bm q}_\mathrm{r} + C(\bm q,\dot{\bm q}) \dot{\bm q}_\mathrm{r}  + G(\bm q) + F(\dot{\bm q}) - \bm \tau.
\end{align*}
Applying (\ref{eq02}) with $\bm v$ $=$ $\dot{\bm q}_\mathrm{r}$ to the above result leads to
\begin{align}\label{eq03a}
M(\bm q) \dot{\mathbf e}_2 = H(\bm q, \dot{\bm q}, \dot{\bm q}_\mathrm{r}, \ddot{\bm q}_\mathrm{r})  + F(\dot{\bm q})  - C(\bm q,\dot{\bm q})\mathbf e_2 - \bm \tau.
\end{align}
It follows from the definitions of $\mathbf e_1$, $\mathbf e_2$ and $\dot{\bm q}_\mathrm{r}$ that $H(\bm q$, $\dot{\bm q}$, $\dot{\bm q}_\mathrm{r}$, $\ddot{\bm q}_\mathrm{r})$ and its corresponding $\Phi_h(\bm q, \dot{\bm q}, \dot{\bm q}_\mathrm{r}, \ddot{\bm q}_\mathrm{r})$ can be denoted as $H(\mathbf x_\mathrm{d}, \mathbf e)$ and $\Phi_h(\mathbf x_\mathrm{d}, \mathbf e)$, respectively. Subtracting and adding $H(\mathbf x_\mathrm{d}, \mathbf 0)$ at the right side of \eqref{eq03a} yields
\begin{align}\label{eq04}
M(\bm q) \dot{\mathbf e}_2 & = \tilde{H}(\mathbf x_\mathrm{d}, \mathbf e) + H(\mathbf x_\mathrm{d}, \mathbf 0)  \notag\\
& + F(\dot{\bm q}) - C(\bm q,\dot{\bm q})\mathbf e_2 - \bm \tau
\end{align}
where $\tilde{H}: \mathbb{R}^{3n}\times\mathbb{R}^{2n}\mapsto\mathbb{R}^{n}$ is given by
\begin{align*}
\tilde{H}(\mathbf x_\mathrm{d}, \mathbf e) := H(\mathbf x_\mathrm{d}, \mathbf e) - H(\mathbf x_\mathrm{d}, \mathbf 0).
\end{align*}
As ${H}$ is of $\mathcal{C}^1$ as in Property 4, there is a globally invertible and strictly increasing function $\rho: \mathbb R^+ \mapsto \mathbb R^+$ so that the following bound condition holds \citep{Xian2004}:
\begin{equation}\label{eq05}
\|\tilde{H}(\mathbf x_\mathrm{d}, \mathbf e)\| \leq \rho(\|\mathbf e\|) \|\mathbf e\|.
\end{equation}
Applying (\ref{eq03}) to $H(\mathbf x_\mathrm{d}, \mathbf 0)$ and using (\ref{eq05a}), (\ref{eq04}) becomes
\begin{align}\label{eq06}
M(\bm q) \dot{\mathbf e}_2 & = \tilde{H}(\mathbf x_\mathrm{d}, \mathbf e) + \Phi_h^T(\mathbf x_\mathrm{d})W_h  \notag\\
& + \Phi^T_f(\dot{\bm q})W_f - C(\bm q,\dot{\bm q})\mathbf e_2 - \bm \tau
\end{align}
where $\Phi^T_h(\mathbf x_\mathrm{d})W_h = H(\mathbf x_\mathrm{d}, \mathbf 0)$ with $\Phi_h: \mathbb R^{3n}\mapsto\mathbb R^{N\times n}$.

Inspired by the human motor learning control mechanism, the control torque $\bm \tau$ is designed as follows:
\begin{align}\label{eq07}
\bm \tau = &\underbrace{K_c \mathbf e} + \underbrace{\Phi^T_h(\mathbf x_\mathrm{d})\hat W_h} + \underbrace{\Phi^T_f(\dot{\bm q}_r)\hat W_f}\\
& \; \bm\tau_{\mathrm{FB}}\;\;\;\;\;\;\;\;\;\; \bm\tau_{\mathrm{FF}} \;\;\;\;\;\;\;\;\;\;\;\;\;\;\; \bm\tau_{\mathrm{FC}}\notag
\end{align}
with $K_c := [I, \Lambda_2]$, in which $\Lambda_2 \in \mathbb R^{n\times n}$ denotes a positive-definite diagonal matrix of control gains, $\bm\tau_{\mathrm{FB}}$ and $\bm\tau_{\mathrm{FF}}$ are PD feedback and adaptive feedforward parts, respectively, and $\bm\tau_{\mathrm{FC}}$ is applied to compensate for the friction $F(\dot{\bm q})$. Replacing $\dot{\bm q}$ by $\dot{\bm q}_\mathrm{r}$ in $\bm\tau_{\mathrm{FC}}$ can make it less noise-sensitive (as $\dot{\bm q}_\mathrm{d}$ and $\mathbf e_1$ ($\dot{\bm q}_\mathrm{r} = \dot{\bm q}_\mathrm{d} + \Lambda_1\mathbf e_1$) are usually less noisy than $\dot{\bm q}$) but still maintains the stability requirement \citep{Slotine1991}.
Substituting (\ref{eq07}) into (\ref{eq06}), one obtains the closed-loop tracking error dynamics
\begin{align}\label{eq08}
M(\bm q) \dot{\mathbf e}_2 & = \tilde{H}(\mathbf x_\mathrm{d}, \mathbf e) + \Phi_h^T(\mathbf x_\mathrm{d})\tilde{W}_h + \Phi^T_f(\dot{\bm q})W_f \notag\\
& - \Phi^T_f(\dot{\bm q}_\mathrm{r})\hat W_f- C(\bm q,\dot{\bm q})\mathbf e_2 - K_c \mathbf e.
\end{align}

\subsection{Composite Error Learning Technique}\label{Learning}

In this subsection, we have $\bm v$ $=$ $\dot{\bm q}$ in \eqref{eq03}.
Applying (\ref{eq03b}) to (\ref{eq01}), one gets a parameterized robot model
\begin{equation}\label{eq09}
\bm\tau(t) = \Phi^T(\bm q(t),\dot{\bm q}(t),\ddot{\bm q}(t))W
\end{equation}
To eliminate the necessity of $\ddot{\bm q}$ in parameter estimation, a linear filter $\frac{\alpha}{s+\alpha}$ is applied to each side of (\ref{eq09}) resulting in
\begin{equation}\label{eq10}
\bm\tau_F(t) = \Phi^T_F(\bm q(t),\dot{\bm q}(t))W
\end{equation}
where $s$ denotes a complex variable, $\alpha \in \mathbb R^+$ is a filtering parameter, $\Phi_F$ $:= \alpha e^{-\alpha t}\ast \Phi$ and $\bm\tau_F := \alpha e^{-\alpha t} \ast \bm\tau$ are filtered counterparts of $\Phi$ and $\bm\tau$, respectively, and ``$\ast$'' is the convolution operator. A predictive model is given by
\begin{equation}\label{eq11}
\hat{\bm\tau}_F(t) = \Phi^T_F(\bm q(t),\dot{\bm q}(t))\hat W(t)
\end{equation}
in which $\hat{\bm\tau}_F \in \mathbb R^n$ is a predicted counterpart of ${\bm\tau}_F$. To facilitate presentation, define an excitation matrix
\begin{equation}\label{eq12}
\Theta(t) := \int_{t-\tau_d}^{t} \Phi_F\big(\bm q(\tau),\dot{\bm q}(\tau)\big) \Phi_F^T\big(\bm q(\tau),\dot{\bm q}(\tau)\big) d\tau.
\end{equation}
Multiplying (\ref{eq10}) by $\Phi_F(\bm q,\dot{\bm q})$, integrating the resulting equality during $[t-\tau_d, t]$ and using (\ref{eq12}), one gets
\begin{equation}\label{eq13}
\Theta(t)W = \int_{t-\tau_d}^{t} \Phi_F\big(\bm q(\tau),\dot{\bm q}(\tau)\big) \bm\tau_F(\tau) d\tau
\end{equation}
which is a filtered, regressor-extended and integrated form of  (\ref{eq09}).
Define a generalized predictive error
\begin{equation}\label{eq14}
\bm\xi(t) := \left\{\begin{array}{l}
\Theta(t)W - \Theta(t)\hat W(t),\; t < T_e\\
\Theta(T_e)W - \Theta(T_e)\hat W(t), \; \mathrm{otherwise}
\end{array}\right.
\end{equation}
where $\Theta W$ is obtainable by (\ref{eq13}). A CEL law with switching $\sigma$-modification is designed as follows:
\begin{gather}\label{eq15}
\dot{\hat{W}} = \gamma\big(\Phi(\mathbf x_\mathrm{d}, \dot{\bm q}_r)\mathbf e_2 + \kappa \bm\xi - \sigma_s \hat{W}\big)
\end{gather}
with $\Phi(\mathbf x_\mathrm{d}, \dot{\bm q}_r) = [\Phi^T_h(\mathbf x_\mathrm{d}), \Phi^T_f(\dot{\bm q}_r)]^T \in \mathbb{R}^{(N+2n)\times n}$, where $\gamma \in \mathbb{R}^+$ is a learning rate, $\kappa \in \mathbb{R}^+$ is a weight factor, and $\sigma_s\hat{W}$ is a $\mathcal{C}^0$ switching ``leaky'' term with
\begin{equation*}
\sigma_s(t) := \left\{\begin{array}{l}
0,\;\;\;\mathrm{if}\;\|{\hat{W}}\| < c_w\\
\sigma_0,\;\mathrm{if}\;\|{\hat{W}}\| > 2c_w\\
\sigma_0\big(\|\hat{W}\|/c_w - 1\big),\;\mathrm{otherwise}
\end{array}\right.
\end{equation*}
and $\sigma_0 \in \mathbb{R}^+$ a constant design parameter. The switching $\sigma$-modification is used to guarantee closed-loop stability with bounded parameter estimation under perturbations if no excitation exists during control  \citep{Ioannou1996}.

\begin{figure}[!t]
\centering
 \includegraphics[width = 3.4in]{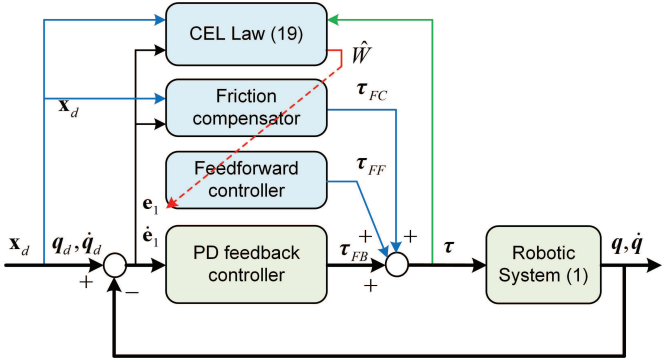}
 \caption{A block diagram of the CEL robot control scheme.}
 \label{Fig03}
 \end{figure}

The overall closed-loop system that combines the tracking error dynamics with the parameter estimation error dynamics is presented according to \eqref{eq08} and \eqref{eq15} as follows:
\begin{align}\label{eq16a}
\left\{\begin{array}{l}
\dot{\mathbf e}_1 = \mathbf e_2 - \Lambda_1\mathbf e_1\\
\dot{\mathbf e}_2 = M^{-1}(\bm q)\big(\tilde{H}(\mathbf x_\mathrm{d}, \mathbf e) - C(\bm q,\dot{\bm q})\mathbf e_2 - K_c \mathbf e\\
\quad\quad\;\, + \Phi_h^T(\mathbf x_\mathrm{d})\tilde{W}_h + \Phi^T_f(\dot{\bm q})W_f - \Phi^T_f(\dot{\bm q}_\mathrm{r})\hat W_f\big)\\
\dot{\tilde{W}} = -\gamma\big(\Phi(\mathbf x_\mathrm{d}, \dot{\bm q}_\mathrm{r})\mathbf e_2 + \kappa \bm\xi - \sigma_s \hat{W}\big)
\end{array}.\right.
\end{align}
A block diagram of the CEL robot control scheme is given in Fig. \eqref{Fig03}.
If there exists $T_e, \sigma_e, \tau_d \in \mathbb R^+$ such that the IE condition $\Theta(T_e) \geq \sigma_e I$ holds, the control parameters $\Lambda_1$, $\Lambda_2$, $\gamma$ and $\sigma_0$ can be properly selected so that the closed-loop system has semiglobal stability in the sense that all signals are of $L_\infty$ and $\mathbf e(t)$ asymptotically converges to \textbf{0} on $t \in [0, \infty)$, and both $\mathbf e(t)$ and $\tilde{W}(t)$ exponentially converge to \textbf{0} on $t \in [T_e, \infty)$.
The above results can be proven based on the Filippov's theory of differential inclusions and the LaSalle-Yoshizawa corollaries for nonsmooth systems \citep{Fischer2013}, where the details are omitted here due to the page limitation.

\section{Industrial Robot Application}\label{Example}

\begin{figure}[!t]
\centering
\includegraphics[width = 3.2in]{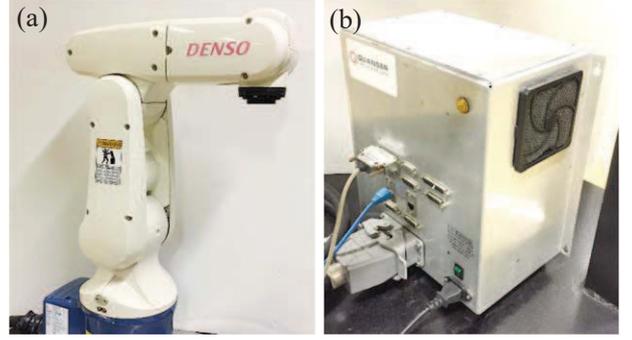}
\caption{Experimental setup: A 6-axis articulated robot. (a) A Denso robot arm (Type: VP6242G). (b) A Quanser open architecture real-time control module.}
\label{FigG1}
\end{figure}

The proposed CEL controller is implemented on a DENSO robot arm with a Quanser real-time control module [see Fig. \ref{FigG1}]. Each joint of the robot arm is driven by an AC servo motor with a speed reducer, where the gear ratios of the three joints used in the experiments are 160, 120, and 100, respectively. A 17-bit absolute rotary encoder is used to measure the angle of each motor. Therefore, the resolutions of the three joints are $3.0\times10^{-7}$rad, $4.0\times10^{-7}$rad and $4.8\times10^{-7}$rad, respectively. The sampling time of the real-time control module is 1 ms.

The robot regression model in \cite{Xin2007} is introduced for implementation. The proposed CEL control law comprised of (\ref{eq07}) and (\ref{eq15})  is rewritten as follows:
\begin{equation*}
\left\{ {
\begin{array}{lcl}
\bm \tau = K_c \mathbf e + \Phi^T(\mathbf x_\mathrm{d}, \dot{\bm q}_r)\hat W\\
\dot{\hat{W}} = \gamma\big(\Phi(\mathbf x_\mathrm{d}, \dot{\bm q}_r)\mathbf e_2 + \kappa \bm\xi - \sigma_s \hat{W}\big)
\end{array}
} \right.
\end{equation*}
where the values of the control parameters are selected as $\Lambda_1 =$ diag(4, 4, 8) and $\Lambda_2 =$ diag(6, 6, 1.5) in \eqref{eq07}, $\alpha $ $=$ 5 in \eqref{eq10}, $\tau_d$ $ =$ 4 in \eqref{eq12}, and $\gamma =$ 0.15, $\kappa =$ 0.5, $\sigma_0 =$ 0.1, $c_w =$ 5 and $\hat{W}(0) = \bm 0$ in (\ref{eq15}).
A baseline controller is chosen as the classical FEL control law as follows:
\begin{equation*}
\left\{ {
\begin{array}{lcl}
\bm \tau = K_c \mathbf e + \Phi^T(\mathbf x_\mathrm{d}, \dot{\bm q}_r)\hat W\\
\dot{\hat{W}} = \gamma\big(\Phi(\mathbf x_\mathrm{d}, \dot{\bm q}_r)\mathbf e_2 - \sigma_s \hat{W}\big)
\end{array}
} \right.
\end{equation*}
where the control parameters are selected to be the same values as the proposed control law for fair comparison.

\begin{table*}[!tb]
  \centering
  \begin{threeparttable}
    \renewcommand{\arraystretch}{1.3}
    \caption{A comparison of performance indices for the two controllers}
    \centering
    \begin{tabularx}{7.1in} {l @{\extracolsep{\fill}} c c c|c c c}
         \toprule \hline
         \multirow{2}*{Ranges of $\mathbf e_1$ and $\bm\tau$} & \multicolumn{3}{c}{The classical FEL Control} & \multicolumn{3}{c}{The proposed CEL control} \\
         \cline{2-7}
                    &Joint 1  & Joint 2 & Joint 3 &   Joint 1 & Joint 2  & Joint 3 \\
         \midrule
        $\mathbf e_1$ before learning ($^\circ$) & [$-$1.266, 2.016] & [$-$1.775, 1.976] & [$-$3.876, 2.297]
        & [$-$0.921, 0.912] & [$-$1.590, 1.911] & [$-$2.973, 0.908] \\
        $\mathbf e_1$ after learning ($^\circ$) & [$-$0.679, 1.044] & [$-$1.519, 0.969] & [$-$3.136, 1.509]
        & [\textbf{$-$0.284, 0.152}] & [\textbf{$-$0.415, 0.098}] & [\textbf{$-$1.708, 1.406}] \\
        $\bm\tau$ before learning (N.m) & [$-$5.478, 11.08] & [$-$11.21, 6.738] & [$-$10.51, 8.289]
        & [$-$1.307, 7.059] & [$-$6.246, 3.358] & [$-$7.832, 5.458] \\
        $\bm\tau$ after learning (N.m) & [$-$5.165, 7.909] & [$-$9.304, 8.865] & [$-$10.90, 6.932]
        & [\textbf{$-$1.837, 6.604}] & [\textbf{$-$7.927, 3.721}] & [\textbf{$-$7.861, 5.774}] \\
        \hline \bottomrule
    \end{tabularx}
  \end{threeparttable}
\end{table*}

To verify the learning ability of the proposed controller, the desired joint position ${\bm q}_\mathrm{d}$ is expected to be simple. Consider a regulation problem with ${\bm q}_\mathrm{d}$ being generated by
\begin{equation*}
\begin{bmatrix}
  \dot q_{\mathrm{d}i}\\
  \ddot q_{\mathrm{d}i}
\end{bmatrix}
= \begin{bmatrix}
  0 &   1 \\
-36 & -12
\end{bmatrix}
\begin{bmatrix}
  q_{\mathrm{d}i}\\
  \dot q_{\mathrm{d}i}
\end{bmatrix}
+ \begin{bmatrix}
  0\\
  36
\end{bmatrix} q_{\mathrm{c}i}
\end{equation*}
with $i =$ 1 to 3 and $\dot{\bm q}_\mathrm{d}(0) =$ \textbf{0}, where $q_{\mathrm{c}i}(t)$ is a step trajectory that repeats every 50 s. The experiments last for 250s, and thus, there are five control tasks. The units of joint position and torque are rad and N.m, respectively. Let $\mathbf e_1 = [e_{11}, e_{12}, e_{13}]^T$, where $e_{1i} \in \mathbb R$ denotes the position tracking error for Joint $i$ with $i =$ 1 to 3.

Table I provides control results of the two controllers for both the first (before learning) and the last tasks (after learning). For the classical FEL control, there exists a large tracking deviation between the desired position $\bm q_\mathrm{d}$ and the actual position $\bm q$ for the first task. After the learning for 200 s, the tracking performance is slightly improved for the last task. As an example, the range of the tracking error $e_{11}$ after learning for Joint 1 is reduced from [$-$1.266, 2.016] (before learning) to [$-$0.679, 1.044] ($\approx$52\% of that before learning) under the classical FEL control.

For the proposed CEL control, a large tracking error $\mathbf e_1$ still exists before learning. During the first task, the IE condition is met. After the learning for 200 s, the proposed CEL control improves significantly in tracking accuracy compared with that before learning. For example, the range of $e_{11}$ after learning for Joint 1 is reduced from [$-$0.921, 0.912] (before learning) to [$-$0.284, 0.152] (only $\approx$24\% of that before learning) under the proposed FEL control. The strong learning capacity of the proposed CEL control is also clearly shown by comparing the ranges of  $\mathbf e_1$ under the two controllers in Table I.

It is also demonstrated in Table I that the maximal control torques $\bm\tau$ of the proposed CEL control are much smaller than those of the classical FEL control for all joints and tasks, which implies that the proposed CEL control is able to achieve much better tracking accuracy even using much smaller control gains and much less energy cost. This is because the improved feedforward control resulting from accurate parameter estimation is beneficial for reducing feedback control torques. The control torque $\bm\tau$ of joint 3 shows slight oscillations compared with those of joints 1 and 2 due to its more significant joint elasticity caused by the unique synchronous belt drive mechanism. This is also the reason why the performance improvement of joint 3 by the proposed CEL control shown in Table I is not as significant as those of joints 1 and 2.

\begin{figure}[!t]
\centering
\includegraphics[width = 3.4in]{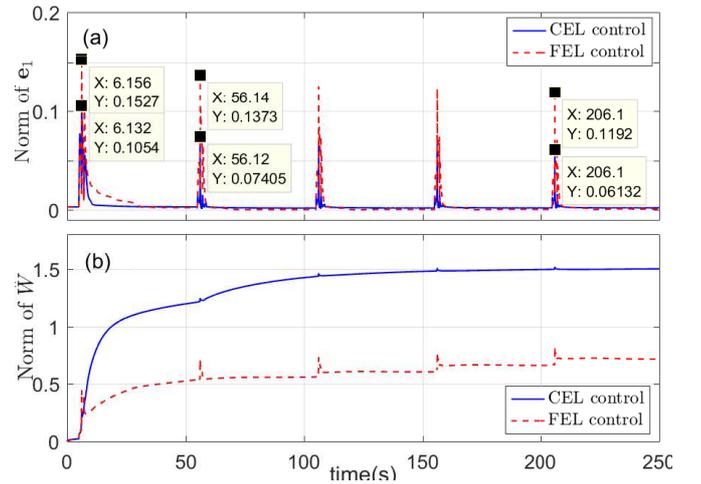}
\caption{A comparison of control trajectories for the two controllers. (a) The norm of the tracking error $\mathbf e$. (b) The norm of the parameter estimate $\hat W$.}
\label{Fig10}
\end{figure}

Fig. \ref{Fig10} provides performance comparisons of the two controllers.
For the classical FEL control, as the PE condition does not hold for the entire control process, no parameter convergence is shown [see Fig. \ref{Fig10}(b)]. In sharp contrast, the proposed CEL control achieves fast convergence of $\|\hat{W}\|$ to a certain constant [see Fig. \ref{Fig10}(b)]. This is consistent with the theoretical analysis: The proposed CEL control only requires the much weaker IE condition for parameter convergence, which can be satisfied during the transient process of the first task.
Also, the CEL control achieves a smaller $\|\mathbf e_1\|$ than the FEL control even from the first control task owing to the predictive error feedback in the CEL law, and maintains the superior tracking performance during the entire control process [see Fig. \ref{Fig10}(a)].

\section{Conclusions}\label{Conclusions}

In this paper, a novel CEL framework has been developed for robot control under discontinuous friction. Compared with the classical FEL control, the distinctive features of the proposed approach include: 1) Semiglobal stability of the closed-loop system is ensured without high feedback gains; 2) exact robot modeling is ensured by the weaken IE condition. The proposed approach has been applied to a DENSO industrial robot, and experimental results have shown that it is superior with respect to tracking accuracy and control energy compared with the classical FEL control. Future work would focus on the optimization of the experimental setup to speed up parameter convergence and the applications of the proposed approach to more real-world robotic systems \citep{Liu2021a, Liu2021b}.

\end{document}